\documentclass[12pt]{article}

\usepackage{palatino}
\usepackage{wrapfig,graphicx,psfrag,xspace,color,latexsym,amsmath,amssymb,booktabs}
\usepackage{comment}
\usepackage{setspace}
\usepackage[dvips]{epsfig}      
\usepackage{longtable}          
\usepackage{url,hyperref}       
\usepackage{multirow}           
\usepackage{color}
\usepackage{algorithm}
\usepackage{algorithmic}
\usepackage{graphicx}
\usepackage{tikz}
\usepackage{xcolor}
\usepackage{euscript,enumerate,float,afterpage}
\usepackage{rotate}
\usepackage{verbatim}
\usepackage{epstopdf}
\usepackage{hhline}
\usepackage{subcaption}
\usepackage[font=footnotesize,labelfont=bf]{caption}
\usepackage[utf8]{inputenc}
\usepackage{listings}
\usepackage{xcolor}

\definecolor{codegreen}{rgb}{0,0.5,0}
\definecolor{codegray}{rgb}{0.5,0.5,0.5}
\definecolor{codepurple}{rgb}{0.58,0,0.82}
\definecolor{backcolour}{rgb}{0.95,0.95,0.92}
\lstdefinestyle{mystyle}{
  backgroundcolor=\color{backcolour},   commentstyle=\color{codegreen},
  keywordstyle=\color{magenta},
  numberstyle=\tiny\color{codegray},
  stringstyle=\color{codepurple},
  basicstyle=\ttfamily\large,
  breakatwhitespace=false,         
  breaklines=true,                 
  captionpos=b,                    
  keepspaces=true,                 
  numbers=left,                    
  numbersep=5pt,                  
  showspaces=false,                
  showstringspaces=false,
  showtabs=false,                  
  tabsize=2,morekeywords={predict,detect},
    keywordstyle=\bfseries,
    showstringspaces=false
}

\newcommand{\squishlist}{
   \begin{list}{$\bullet$}
    { \setlength{\itemsep}{0pt}      \setlength{\parsep}{0pt}
      \setlength{\topsep}{3pt}       \setlength{\partopsep}{0pt}
      \setlength{\listparindent}{-2pt}
      \setlength{\itemindent}{-5pt}
      \setlength{\leftmargin}{1em} \setlength{\labelwidth}{0em}
      \setlength{\labelsep}{0.5em} } }

\newcommand{\squishend}{
    \end{list}  }

\begin{document}

\title{Robust and Active Learning for\\
Deep Neural Network Regression}

\author{
Xi Li, George Kesidis, David J. Miller \\
Pennsylvania State University \\
\{xzl45,gik2,djm25\}@psu.edu \\
~\\
\begin{tabular}{cc}
Maxime Bergeron,  Ryan Ferguson  & Vladimir Lucic\\
Riskfuel & Imperial College \\
\{mb,rf\}@riskfuel.com & v.lucic@imperial.ac.uk
\end{tabular}
}

\date{}

\maketitle


\section{Introduction}

We describe a heuristic method for active learning of a regression model, in particular of a deep neural network model,
e.g. \cite{TPS16,Kading18,WLH18,LMPH19,Ren20}.
In the following, an automated ``oracle" $Z$ capable of providing 
real-valued supervision (a regression target) for samples, for
purposes of training, is assumed available. 
A parameterized model $Y$ for regression, which could be a deep neural network (DNN),
is typically trained to approximate the
oracle based on the mean squared error (MSE) training objective, i.e.,
\begin{align} \label{MSE0}
  \frac{1}{|S|} \sum_{x\in S} \left|Y(x)-Z(x)\right|^2
\end{align}
where $S$ is the training set of supervised input samples $x$, with the $Z(x)$ the supervising target for $x$.
The benefit of a
DNN model is that it can perform inference at much higher speed 
than the oracle.
This benefit is herein assumed to greatly outweigh the cost of invoking
the oracle during training.
Moreover, a DNN model is preferable to simpler alternatives
when the oracle is a complicated (highly nonconvex) function of 
the input and the inputs belong to a 
high-dimensional sample space.

The following describes an active learning approach to this training problem.
We give an approach that iteratively enriches the 
training set with new supervising examples, which inform the learning and result in improved accuracy of the (re-)learned model.

\section{Overview of the Method}\label{sec:method}

Initially, the DNN is trained based on a random sampling of
the input space $S_0$ (where each sample is supervised by the oracle).
Seeded by the initial training samples that
exhibit the largest absolute 
errors $|Y-Z|$, gradient-ascent search is then used to identify a set of local maximizers, $M_0$, of the squared error, $|Y-Z|^2$,
where a finite-difference approximation is
used for the gradient of $Z$ and the gradient of $Y$ is
directly computed by back-propagation with respect to the
input variables, e.g., \cite{Hwang97}.
As often done for gradient based optimization for training $Y$ (deep learning),
here the step size can be periodically reduced 
by a fixed proportion.
Once the set of local maximizers are identified, amongst any subset that are all very highly proximal to each other, only one need be retained.

In summary, a gradient-ascent sequence with index $n=0,1,2,...$, seeking a local maximizer of the square error ${\mathcal E}=|Y-Z|$, is: 
$$x_{n+1} = x_n + s_n\nabla{\mathcal E}^2(x_n), $$
where $x_0$ is the initialization, the step size $s_n$ is non-increasing with $n$, a
finite-difference approximation is used for the gradient of the oracle $Z$, and the gradient of the neural network model $Y$ is computed by back-propagation.

Using the training set $S_1=S_0\cup M_0$,
the DNN is then retrained in the manner of active learning.
Depending on the application,
 a differently weighted combination of error terms can be used
as the training objective, e.g.,
\begin{align} \label{MSE-weighted}
  \alpha \frac{1}{|S_0|} \sum_{x\in S_0} \left|Y(x)-Z(x)\right|^2
+(1-\alpha)\frac{1}{|M_0|} \sum_{x\in M_0} \left|Y(x)-Z(x)\right|^2
\end{align}
for some $0 < \alpha < 1$.
If $S_0\cap M_0=\emptyset$
and $\alpha = |S_0|/|S_1|$,   then this objective
is just \eqref{MSE0} with $S=S_1$.
Depending on the application, one might want to give
greater weight to the local maximizers in subsequent learning iterations, 
i.e., by taking $\alpha < |S_0|/|S_1|$.

The foregoing process is iteratively repeated
in the manner of classical iterated gradient based
hill-climbing  exploration/exploitation optimization 
methods, more recently called
Cuckoo search \cite{YD09}:
At step $k+1$,  gradient ascent search  
of the error $|Y-Z|^2$ is seeded with the 
elements of $S_k$ with largest absolute errors,
and uses smaller initial step size and
tighter stopping condition than for step $k$.

Many obvious variations of the foregoing approach are possible.
Though we assume the frequent use of the
oracle is justified for training,
computational costs of training could be much more significant for
more complex oracles. 
Given this, note the likely  significant
computational advantage of the use of gradient ascent search over ``pure"
iterated random sampling of regions of sample space with
higher error, i.e.,  a kind of iterative importance sampling
based Monte Carlo.

\section{Discussion: Overfitting and regularization}

Typically, one has no idea {\it a priori} how large a model is needed for a given application domain.
So a particular DNN may or may not be (initially) overparameterized
with respect to its training set.
Low accuracy on the training set may indicate too few parameters
(insufficient DNN capacity to learn).
On the other hand,
low accuracy on a supervised validation set held out from training 
(poor generalization performance)
with high accuracy on the training set may indicate too many parameters
(overfitting to the training set).

Suppose that in each iteration of the training method described in Section
\ref{sec:method}, 
a feed-forward neural network with five fully connected internal layers
and 256 neurons per layer (alternatively, we could use ``convolutional"
layers with far fewer parameters per layer)
is initially used.
Note that as local maximizers of error are identified, it is possible that
the resulting regression problem becomes more complex, particularly if
the number and density of local extrema increases. Thus, this
DNN may need to be ``regularized" initially to avoid
overfitting (e.g., using dropout when training),  while
in later iterations the DNN may not have sufficient 
capacity and so the number of layers/neurons may need to be
increased to achieve required accuracy on
the validation set.

\section{Experimental Results}

Some experimental results are now given for a simple oracle used
to value a single-barrier option.
The input is $5$ dimensional:
barrier over spot price, strike over spot price, 
time to maturity, volatility,  and
interest rate.
Each input sample was confined to a ``realistic" interval and normalized for training.


The initial training set had size
$|S_0|=200$k and there is a test set of $10$k samples, the
latter used to evaluate accuracy.
These sets were taken by selecting sample $5$-tuples uniformly
at random and discarding those which were  extraneous.

Under PyTorch,
we used a feed-forward DNN with five fully
connected internal layers each having 512 ReLU neurons. The learning rate (step size) was 0.01 initially and divided by 10 every 50 epochs.
Training of the DNN halted when the normalized change in
training MSE was less than $0.001$ over 10 epochs.
Dropout \cite{Ruder16} was not  employed when training.

To identify the local maximizers of the squared error of
the DNN trained on $S_0$,
gradient ascent is performed on the square error
starting from an initial point with large error, with the 
step size divided by $10$ every $30$ epochs, with
initial step size $0.001$ and with the
termination condition when the squared error is $0.001$.
To approximate the gradient of the oracle $Z$, a
first-order finite difference with parameter 
$0.001$ was used. 
Local maximizers were deemed identical when 
the Euclidean norm between them was less than $0.001$. 
$|M_0|=10$k unique local maximizers were thus found by
seeding gradient ascent with the $5$\% of samples of $S_0$ 
having highest absolute error. 

From Table \ref{tab:tab1} consider $\alpha=1$ (the DNN trained
on $S_0$).
The training set MSE is $0.0646$ (dollars)
and the test set MSE is $0.0688$, i.e., the performance
on the training set ``generalizes" well to the test set.
Also,
the mean absolute error (MAE) on the test set
is 0.1734 (dollars), while the MAE on the 
set of local maximizers of square error, $M_0$, is 1.168.
That is, this DNN is not extremely accurate, particularly on
the local maximizers.

\begin{table}
\centering
\resizebox{\columnwidth}{!}{
\begin{tabular}{ l|lllllllllll } 
\toprule
$\alpha$ & 1 & 0.99 & 0.98 & 0.97 & 0.96 & 0.95 & 0.94 & 0.93 & 0.92 & 0.91 & 0.9 \\
\hline
Training MSE & 0.0646 & 0.0701 & 0.0463 & 0.0425 & 0.0506 & \textbf{0.0359} & 0.0448 & 0.0422 & 0.0411 & 0.0435 & 0.0475\\
Test MSE & 0.0688 & 0.07 & 0.0501 & 0.0495 & 0.0513 & \textbf{0.0381} & 0.0437 & 0.0486 & 0.0452 & 0.0521 & 0.046 \\
Maximizer MSE & 1.557 & 2.916 & 2.981 & 0.8214 & 2.196 & \textbf{0.7501} & 1.012 & 0.8615 & 3.262 & 1.153 & 1.493 \\
Training MAE & 0.1686 & 0.1777 & 0.1451 & 0.1413 & 0.1524 & \textbf{0.1298} & 0.1436 & 0.1422 & 0.1373 & 0.1437 & 0.1435 \\
Test MAE & 0.1734 & 0.1696 & 0.1419 & 0.1408 & 0.1501 & \textbf{0.1301} & 0.1425 & 0.1417 & 0.1375 & 0.1457 & 0.1412 \\
Maximizer MAE & 1.168 & 0.9941 & 0.8234 & 0.7524 & 0.8511 & \textbf{0.6821} & 0.7345 & 0.7287 & 0.7557 & 0.7477 & 0.8271 \\
\bottomrule
\end{tabular}
}
\caption{Results for the DNN trained on $S_0$ ($\alpha=1$) and $S_1=S_0\cup M_0$ ($\alpha < 1$).}\label{tab:tab1}
\end{table}

Retraining the DNN\footnote{Alternatively, the model
based on $S_0$ could be fine-tuned \cite{LMPH19}.}
using $S_1=S_0\cup M_0$  was
terminated when the normalized change in the training MSE over 10 iterations
was less than $0.001$.
To find $10$k of its local maximizers with respect to squared error,
gradient ascent was used, 
reducing the step size by $10$ every $30$ iterations, with
initial step size $0.00001$ 
and stopping condition when the normalized changed in square error
$< 0.00001$. The seeds for gradient ascent are the $5\%$ of samples of $S_1$ 
having highest absolute error. 

From Table \ref{tab:tab1}, for $\alpha <1$
(the DNN trained on $S_1=S_0\cup M_0$),
note that $\alpha = |S_0|/|S_1| = 0.95$ means that all
samples (those in $S_0$ and $M_0$) are weighted equally.
Compared to the original DNN, this DNN has lower
training and test MSE and lower test and maximizer MAEs, 
even though the learning task is more difficult (given the additional $|M_0|$ training samples and the fact that
the DNN architecture (its model size) has not been changed).

We notice that, compared with the case with $\alpha=1$, the maximizers in some cases (e.g., $\alpha=0.99$) have higher MSE but lower MAE. For $\alpha=0.99$, with the $0.1\%$ maximizers having highest absolute errors removed, the MSE and MAE are $1.361$ and $0.9671$, respectively, and are smaller than the MSE and MAE for $\alpha=1$. Hence, the inconsistency between maximizer MSE and MAE is caused by the extreme instances, which significantly lift the MSE.

The foregoing retraining process can be repeated, and terminated
when no new local maximizers are found, and when the increment in generalization 
performance on both the test set and the set of local maximizers
$\cup_{k\geq 0} M_k$ levels off without indication that the
learning capacity of the DNN has been reached.
\pagebreak

\bibliographystyle{plain}

\end{document}